\documentclass[twocolumn]{svjour3} 
\smartqed 

\usepackage{graphicx}%
\usepackage{multirow}%
\usepackage{amsmath,amssymb,amsfonts}%
\usepackage{xcolor}%
\usepackage{booktabs}%
\usepackage{algorithm}%
\usepackage{algorithmicx}%
\usepackage{algpseudocode}%
\usepackage{flushend}

\usepackage{xcolor}
\usepackage{hyperref}
\usepackage{setspace}
\usepackage{colortbl}
\usepackage{pifont}

\makeatletter
\def\eg{\emph{e.g.}}

\definecolor{amaranth}{rgb}{0.9, 0.17, 0.31}
\definecolor{applegreen}{rgb}{0.55, 0.71, 0.0}
\definecolor{amethyst}{rgb}{0.6, 0.4, 0.8}
\definecolor{ao}{rgb}{0.0, 0.5, 0.0}
\definecolor{alizarin}{rgb}{0.82, 0.1, 0.26}
\newcolumntype{a}{>{\columncolor{blue!10}}c}

\usepackage{bbm}
\usepackage{bm}
\usepackage[numbers]{natbib}

\begin{document}
\sloppy

\title{
Harnessing Vision-Language Models for Federated Unsupervised Visual Recognition
}

\author{Jian Liang $^{1,2}$
\and Kuangpu Guo $^{3}$
\and Ran He $^{1,2}$
\and Tieniu Tan $^{1,2,4}$ }

\institute{
	$^1$ NLPR \& MAIS, Institute of Automation, Chinese Academy of Sciences (CASIA), China \\
	$^2$ School of Artificial Intelligence, University of Chinese Academy of Sciences, China \\
	$^3$ University of Science and Technology of China\\
	$^4$ Nanjing University, China
	\\
 Corresponding author: Jian Liang \tt{(liangjian92@gmail.com)}
}

\date{Received: August 16, 2026}

\maketitle

\begin{abstract}
Federated learning (FL) is a promising research paradigm that enables collaborative model training across distributed clients without data sharing.
To handle unlabeled data in local clients, existing unsupervised FL research has primarily focused on tasks like clustering and self-supervised representation learning, leaving the critical problem of unsupervised visual recognition largely unexplored.
To bridge this gap, we introduce FedUR, the first framework designed for unsupervised recognition by leveraging the zero-shot capabilities of pre-trained vision-language models like CLIP.
Specifically, FedUR addresses data heterogeneity through a pseudo-labeling mechanism with a class-aware data selection strategy.
This strategy first estimates a global pseudo-label distribution, then allocates pseudo-labeled data to clients accordingly. 
To handle the uncertainty of conventional positive pseudo-labels, FedUR further incorporates class weights and a negative pseudo-labeling strategy.
Furthermore, FedUR proposes a personalized prompt aggregation protocol where general visual prompts are aggregated on the server, and personalized text prompts encoding local knowledge are retained on clients.
Extensive experiments demonstrate that FedUR consistently outperforms baseline methods under both label-level and feature-level distribution shifts.
We believe this work paves the way for further advances in federated unsupervised recognition.
Code will be available at \texttt{\href{https://github.com/tim-learn/FedUR}{github.com/tim-learn/FedUR/}}.
\end{abstract}

\keywords{federated unsupervised learning; vision-language models; personalized prompt aggregation; pseudo-labeling; data heterogeneity}

\section{Introduction}
Federated learning (FL)~\cite{sheller2020federated,li2020federated} has emerged as a pivotal paradigm for decentralized machine learning since its introduction by Google in 2016~\cite{mcmahan2017communication}. 
By enabling collaborative model training without the need to share raw local data, FL offers a compelling privacy-preserving solution that has garnered substantial research interest~\cite{yang2019federated,kairouz2021advances,li2022federated,shi2023understanding}. 
To date, the vast majority of FL research has concentrated on supervised and semi-supervised settings, achieving notable success in tasks such as visual classification~\cite{hsu2020federated,luo2021no,oh2022fedbabu,guo2025exploring}, semantic segmentation~\cite{miao2023fedseg,dong2023federated,paranjape2024federated}, and object detection~\cite{kim2023navigating,jia2024adaptive}.
In contrast, considerably less attention has been paid to the challenging scenario of federated unsupervised learning (FedUSL)~\cite{tian2024towards,kim2023protofl,han2022fedx,yan2024lightweight}, where clients possess only unlabeled data.
Existing FedUSL approaches~\cite{zhang2023federated,pourpanah2025federated,liao2024rethinking} predominantly focus on self-supervised representation learning for downstream tasks or clustering for data grouping. 
However, these methods cannot address zero-shot classification, as they are fundamentally constrained by the absence of labeled supervisory signals.

Vision-language models (VLMs)~\cite{li2019visualbert,alayrac2022flamingo,zhai2023sigmoid}, and particularly contrastive language-image pre-training (CLIP)~\cite{radford2021learning}, have demonstrated remarkable representation and zero-shot generalization capabilities~\cite{koh2023generating,faysse2024colpali,marcu2024lingoqa,zhang2024vision,dong2026adapting}.
By learning a shared embedding space from large-scale image-text pairs, CLIP enables powerful zero-shot classification and alleviates the dependency on labeled data, paving the way for unsupervised learning methods~~\cite{huang2022unsupervised,tanwisuth2023pouf,abdelfattah2023cdul,liang2024realistic,wang2024hard}. 
This capability presents a compelling opportunity to address classification tasks within the FedUSL framework, an avenue that remains largely unexplored.

In this paper, we pioneer the extension of FedUSL to classification by leveraging CLIP, with each client initialized with the same pre-trained CLIP model and holding only unlabeled local data.
Following standard FL protocols~\cite{xie2024perada,yang2024fedas}, clients periodically upload non-data knowledge, such as statistical information or model parameters, to a central server for aggregation and receive updated global knowledge in return.
A natural approach within this framework is to use CLIP's zero-shot predictions as pseudo-labels to supervise local training. 
However, the inherent bias of CLIP and the unknown, potentially non-IID label distribution across clients~\cite{huang2022unsupervised,menghini2023enhancing,luo2021no} can lead to globally imbalanced training across categories. 
This label skew causes model updates to diverge among clients, rendering conventional aggregation methods ineffective.

To address these challenges, we propose \textbf{FedUR}, a novel method with three core components: cooperative data selection, dual pseudo-labeling, and personalized prompt aggregation.
The cooperative data selection mechanism selects samples with a balanced pseudo-label distribution across all classes, mitigating the bias inherent in CLIP and the unknown client data distribution.
Specifically, each client first applies confidence- and entropy-based filtering to estimate its local pseudo-label distribution, which is then transmitted to the server.
The server subsequently rebalances each client's distribution by allocating per-class pseudo-label counts according to global category-wise proportions. 
This rebalancing corrects for label skew across clients; however, class bias may persist locally. 
To address this residual bias, we employ a class-weighted cross-entropy loss that penalizes overrepresented classes during local training.
Moreover, we introduce negative pseudo-labels~\cite{rizve2021defense} on selected samples to suppress the probability mass assigned to the most unlikely categories, further strengthening the quality of the pseudo-labels.

Beyond pseudo-labeling, we introduce a personalized prompt aggregation strategy in FedUR to balance global collaboration and local personalization. 
For communication and computational efficiency, we optimize only the visual and textual prompts while freezing the backbone. 
Prior work~\cite{xing2023dual} has shown that visual prompts capture general feature representations, whereas textual prompts focus on category-specific information.
Aggregating prompts from both modalities may introduce parameter conflicts, particularly under label skew.
We therefore transmit only visual prompts to the server for aggregation, while retaining textual prompts locally. 
This design fosters global collaboration through shared visual prompts and supports personalization through client-specific textual prompts that better reflect each client's label distribution. 
Extensive experiments on standard federated benchmarks under Dirichlet-based, quantity-based, and feature skews validate FedUR's effectiveness against baseline methods.
Our main contributions are summarized as follows:

\begin{itemize}
\item We pioneer the extension of federated unsupervised learning to classification tasks by leveraging CLIP, an underexplored direction in prior work.
\item We propose FedUR to mitigate CLIP's inherent bias through a cooperative pseudo-labeling strategy that constructs balanced pseudo-labels, complemented by a class-weighted cross-entropy loss and negative pseudo-labeling for refined local training.
\item We introduce a personalized prompt aggregation protocol in FedUR that enables effective global collaboration while preserving local personalization.
\item Extensive experiments on standard datasets under both label and feature skews demonstrate the effectiveness of FedUR against baseline methods.
\end{itemize}

This paper extends our prior conference work~\cite{guo2025cooperative} in four major respects: 
(1) we identify residual class bias during local training and design a class-weighted objective to mitigate it; 
(2) we introduce negative pseudo-labels to suppress probabilities on unlikely classes, further refining pseudo-label quality; 
(3) we broaden the experimental evaluation with two additional datasets (DomainNet and Office-Home) and more non-IID settings beyond label skew; and 
(4) we conduct a thorough analysis of the proposed framework, focusing on the newly incorporated components.

\section{Related Work}
\subsection{Federated Learning for VLMs}
Vision-language model fine-tuning has recently been incorporated into the federated framework to mitigate computational costs while addressing data scarcity and heterogeneity~\cite{qiu2023text,li2024position,halbe2023hepco,su2022cross}.
Early efforts, notably PromptFL~\cite{guo2023promptfl} and FedCLIP~\cite{lu2023fedclip}, directly apply standard CLIP fine-tuning to the federated setting and achieve competitive performance. 
Both adopt parameter-efficient approaches but differ in implementation: PromptFL fine-tunes continuous prompts, while FedCLIP employs an attention-based adapter for the image encoder.
To better model heterogeneity across clients, subsequent work has introduced various personalization strategies. pFedPrompt~\cite{guo2023pfedprompt} develops a non-parametric approach that equips each client with a personalized attention module. 
pFedPG~\cite{yang2023efficient} generates personalized prompts based on client visual prompts to better align with local data distributions. 
FedOTP~\cite{li2024global} leverages optimal transport to balance global consensus and local personalization through a strategic combination of global and local prompts. 
pFedMoAP~\cite{luo2025mixture} personalizes prompt learning via a mixture of experts, allowing each client to adaptively combine multiple downloaded prompts for local alignment. 
PromptFolio~\cite{pan2024federated} theoretically analyzes prompt-based FL through feature learning, revealing that an optimally mixed global-local prompt portfolio balances generalization and personalization.
Despite these advances, prior work has predominantly focused on supervised federated learning with VLMs. 
In this paper, we instead explore leveraging VLMs for federated unsupervised classification, capitalizing on their zero-shot capabilities.

\subsection{Unsupervised Learning for VLMs}
Despite advances in fine-tuning VLMs, high annotation costs remain a major bottleneck for supervised learning in real-world visual applications, motivating the use of unlabeled data for downstream tasks~\cite{dong2026adapting}. 
Recent years have witnessed considerable efforts in this direction, particularly through pseudo-labeling strategies~\cite{huang2022unsupervised,menghini2023enhancing,tanwisuth2023pouf,zhang2024candidate,jia2024mltu}. 
UPL~\cite{huang2022unsupervised} and FPL~\cite{menghini2023enhancing} enforce balanced pseudo-label selection across categories, while CPL~\cite{zhang2024candidate} generates multiple candidate labels per instance to improve accuracy. 
In contrast, GDA-U~\cite{wang2024hard} models data with a Gaussian mixture and employs the EM algorithm for closed-form classification, whereas UEO~\cite{liang2024realistic} adopts a confidence-aware entropy minimization strategy.
In parallel, test-time prompt tuning has emerged as a popular paradigm for unsupervised VLM adaptation~\cite{sheng2025illusion,dong2026adapting}, especially for a single unlabeled image.
TPT~\cite{shu2022test} and its extensions~\cite{yoon2024c,sheng2025r} minimize marginal entropy to update text prompts for individual test samples, while training-free counterparts~\cite{zanella2024test,farina2024frustratingly,jiang2026use} are developed to avoid backpropagation. 
Moreover, TDA~\cite{karmanov2024efficient} further tackles adaptation under a stream of unlabeled test samples.
Despite their effectiveness, these strategies face substantial challenges when directly applied to decentralized federated unsupervised settings, particularly under data heterogeneity. 
To address this, we propose a cooperative pseudo-labeling strategy that mitigates the inherent biases of pre-trained VLMs and substantially improves pseudo-label quality.

\subsection{Federated Unsupervised Learning}
Federated unsupervised learning (FedUSL) aims to collaboratively train a global model across distributed clients with unlabeled data without raw data sharing. 
Existing FedUSL methods can be broadly categorized into two main lines of work: self-supervised representation learning~\cite{van2020towards,lubana2022orchestra,zhang2023federated,liao2024rethinking,rehman2023dawa} and federated clustering~\cite{stallmann2022towards,dennis2021heterogeneity,ding2026federated}.
For representation learning, FedCA~\cite{zhang2023federated} employs contrastive aggregation with a shared dictionary, while FedU~\cite{zhuang2021collaborative} uploads only client encoders to the server.
FedEMA~\cite{zhuang2022divergence} addresses non-IID data via adaptive EMA updates of the global model, while FedX~\cite{han2022fedx} refines local training through two-sided knowledge distillation.
Recent work~\cite{liao2024rethinking,chen2026understanding} emphasizes alignment regularization to ensure representation consistency across clients.
For clustering, $k$-FED~\cite{dennis2021heterogeneity} proposes a one-shot federated k-means scheme based on Lloyd's method. 
FKM~\cite{garst2024federated} relaxes the assumption of equal local cluster counts and handles a variable number of local clusters.
Subsequent studies~\cite{feng2024federated,liu2026communication} consider federated multi-view clustering, enabling multiple clients to build a global model with distributed multi-view data.
Beyond these two setups, FedGaLA~\cite{pourpanah2025federated} performs client-level gradient alignment to encourage domain-invariant feature learning for unsupervised domain generalization.
The most closely related work to ours is FedUL~\cite{lu2022federated}, which studies an unsupervised scenario where class priors for each client dataset are assumed to be available.
Its core idea is to transform unlabeled client data and modify the client model to construct a surrogate supervised task. 
In contrast, our approach leverages CLIP's zero-shot capability to generate pseudo-labels without requiring such label priors or additional assumptions on the unlabeled data.

\section{Methodology}
\begin{figure*}[tbp]
\centering
\includegraphics[width=0.99\textwidth, trim=2.5cm 3.5cm 3cm 2.5cm, clip]{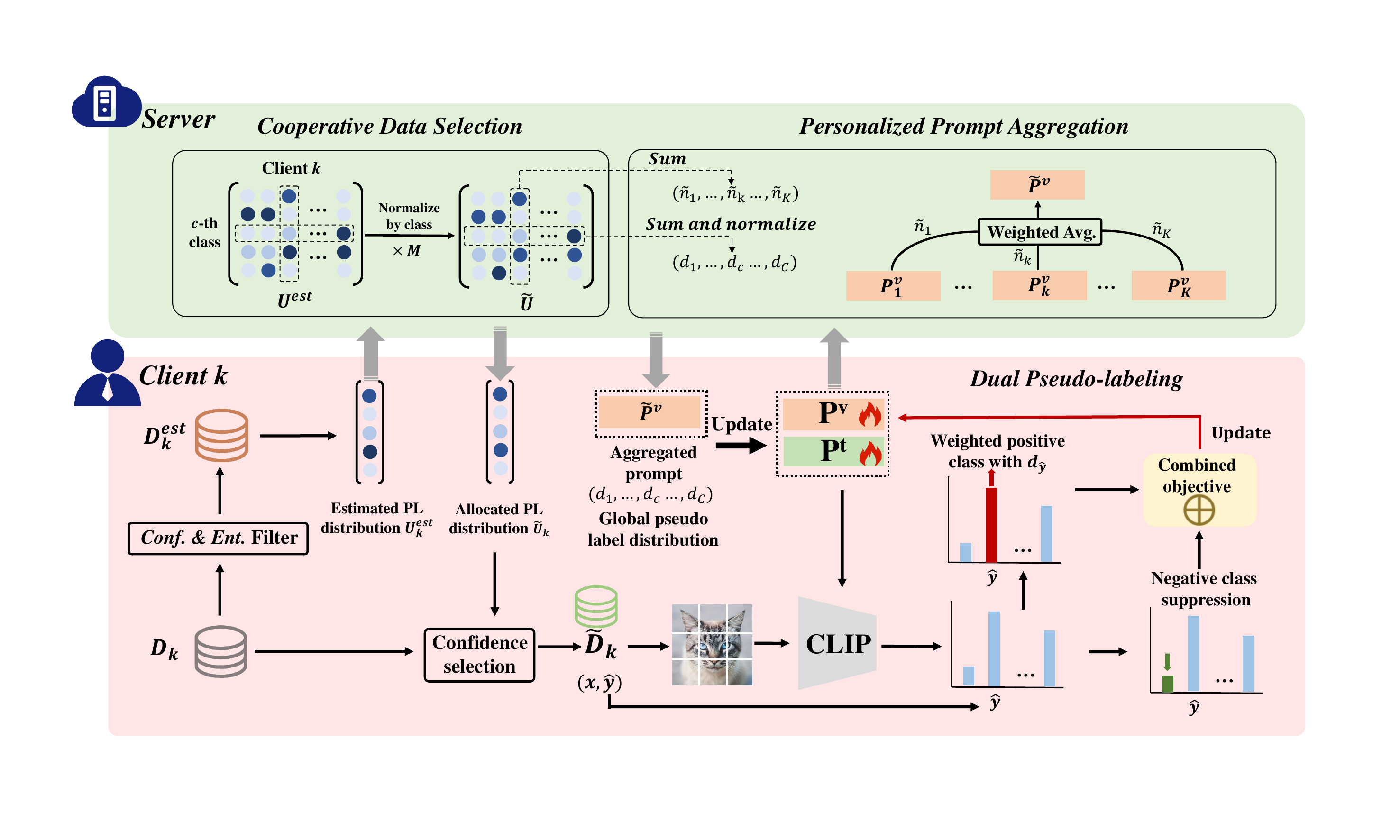}
\caption{
\textbf{Overview of the proposed FedUR framework.} 
The framework begins by filtering unlabeled samples to estimate local label distributions, which are then uploaded to the server. 
The server globally allocates $M$ pseudo-labels per category based on these estimates. 
Clients subsequently perform local training with a dual pseudo-labeling strategy to fully leverage CLIP's predictions. 
To handle data heterogeneity, only visual prompts are aggregated on the server, while textual prompts are retained locally, as they exhibit larger cross-client divergence and better support personalization.} 
\label{fig:framework}
\end{figure*}

In this section, we formulate the problem setup for federated unsupervised recognition and present our proposed method, as shown in Fig.~\ref{fig:framework}. 
Sec.~\ref{sec:pre} reviews a representative VLM, CLIP~\cite{radford2021learning}, along with prompt tuning~\cite{lester2021power}, a popular parameter-efficient fine-tuning technique. 
Sec.~\ref{sec:problem} then outlines the federated unsupervised recognition framework with VLMs and discusses its associated challenges. 
Finally, we introduce FedUR, which builds upon three key strategies: server-side cooperative data selection (Sec.~\ref{sec:coper}), client-side local self-training (Sec.~\ref{sec:local}), and server-side personalized prompt aggregation (Sec.~\ref{sec:prompt}).
The detailed pseudocode of FedUR is presented in Algorithm~\ref{alg:fedur}.

\subsection{Preliminary}
\label{sec:pre}
Visual recognition studies with VLMs have progressed significantly since the development of CLIP~\cite{radford2021learning,zhang2024vision}; we thus adopt CLIP as the default VLM for unsupervised visual recognition.
CLIP employs a dual-branch architecture comprising an image encoder $f^v(\cdot)$ and a text encoder $f^t(\cdot)$, each processing data from its respective modality. 
For zero-shot classification, CLIP uses a hand-crafted prompt template (\eg, ``\texttt{a photo of a [CLASS]}") for each class.
Consider a $C$-way classification task; the text encoder generates class-specific embeddings $\{f^t_c\}_{c=1}^C$, while the image encoder produces a visual embedding $f^v(x)$ for a test image $x$.
The probability ($p_c(x) = p(y=c|x)$) that $x$ belongs to class $c$ is computed via softmax over cosine similarities:
\begin{equation}
p(y=c|x) = \frac{\exp(\text{cos}(f^v(x), f^t_c)) / \tau)}{\sum_{j=1}^{C} \exp(\text{cos}(f^v(x), f^t_j)) / \tau)},
\end{equation}
where $\text{cos}(\cdot,\cdot)$ denotes cosine similarity and $\tau$ is a fixed temperature parameter (\eg, $\tau=0.01$).

Inspired by prompt learning~\cite{liu2023pre} in natural language processing, numerous prompt learning methods have been proposed to adapt VLMs to downstream tasks without full fine-tuning. 
These approaches generally follow three paradigms~\cite{zhang2024vision}: text prompt tuning, visual prompt tuning, and multi-modal prompt tuning.
For example, CoOp~\cite{zhou2022learning} expands a class token \texttt{[CLASS]} into a learnable sentence ``$\texttt{[V]}_1,\texttt{[V]}_2,\dots,\texttt{[V]}_m,\texttt{[CLASS]}$" and optimizes the text prompts $\texttt{[V]}$ via classification loss. 
VPT~\cite{jia2022visual} introduces continuous visual prompts as learnable embeddings in the Transformer's input space.
MaPLe~\cite{khattak2023maple} conducts multi-modal prompt tuning by aligning visual prompts with textual prompts, enabling mutual enhancement between modalities.
For simplicity, we denote the learnable textual and visual prompts as 
$P^t$ and $P^v$, respectively, which are inserted into their corresponding encoders.

\subsection{Problem Formulation}
\label{sec:problem}
The powerful zero-shot classification capability of CLIP reduces reliance on labeled data, enabling classification tasks in FedUSL that were previously intractable. 
In this paper, we extend the FedUSL paradigm to classification by leveraging VLMs, particularly CLIP~\cite{radford2021learning}. 
Each client is initialized with the same pre-trained CLIP model and possesses an unlabeled local dataset $\mathcal{D}_k$ of size $n_k$. 
Clients are also provided with the full set of class names for the task.
Following standard FL protocols~\cite{li2020federated,kairouz2021advances}, clients periodically upload non-data knowledge (\eg, statistical information, model parameters) to the server for aggregation and download updated global information. 
Besides, we focus on FedUSL under data heterogeneity, a common characteristic of real-world datasets, where data distributions in different clients are usually not independently and identically distributed (non-IID).

The primary challenge of federated unsupervised recognition stems from two factors: the inherent bias of CLIP and data heterogeneity. 
On one hand, the pseudo-labels generated by CLIP can be noisy due to its intrinsic bias, which undermines the reliability of unsupervised learning.
On the other hand, data heterogeneity~\cite{li2022federated}, manifesting as label skew and feature skew across clients, further complicates the training process.
Label skew refers to the variation of label marginal distributions $p_i(y)$ across clients, while $p_i(x|y) = p_j(x|y)$ for all clients $i$ and $j$. 
For instance, although the union of all client datasets $\bigcup_{k \in K} \mathcal{D}_k$ covers all categories, each local dataset $\mathcal{D}_k$ may contain only a subset of classes.
Feature skew refers to the variation of input feature marginal distributions $p_i(x)$ across clients, while $p_i(y|x) = p_j(y|x)$ for all clients $i$ and $j$. 
For example, client data may originate from different domains with distinct visual styles.
These heterogeneity-induced issues, together with the noisy pseudo-labels from CLIP, lead to imbalanced training across both categories and clients, causing model updates to diverge and ultimately hindering effective collaboration.

\subsection{Cooperative Data Selection}
\label{sec:coper}
Given CLIP's inherent bias and the label skew across clients, directly applying conventional pseudo-labeling methods leads to unbalanced and poorly representative pseudo-labels. 
Specifically, these methods tend to favor categories preferred by CLIP, even when the resulting predictions are incorrect~\cite{menghini2023enhancing,zhang2024candidate}. 
To address this, we propose a cooperative data selection strategy that promotes a more representative and balanced data distribution across all classes.

Specifically, for each client $k$, we filter samples from the unlabeled dataset $\mathcal{D}_k$ based on prediction confidence and entropy~\cite{huang2022unsupervised,shu2022test} to construct the estimated set $\mathcal{D}^\text{est}_k$ in the following:
\begin{equation}
\mathcal{D}^\text{est}_k = \{(x, \hat{y}) \mid \max_{c} p_c(x) > \tau_1,\; \texttt{Ent}(p(x)) < \tau_2 \},
\end{equation}
where $\hat{y} = \arg\max_c p_c(x)$ is the predicted pseudo-label and $\texttt{Ent}(\cdot)$ denotes Shannon entropy.
The thresholds $\tau_1$ and $\tau_2$ are set to the medians of prediction confidences and entropies over the training set, respectively.
Adopting these dataset-level medians eliminates the need for per-dataset hyperparameter tuning, making our filtering process highly adaptive.
However, due to CLIP's inherent bias, $\mathcal{D}^\text{est}_k$ tends to be dominated by samples from CLIP-preferred categories, even when predictions are incorrect~\cite{menghini2023enhancing,zhang2024candidate}. 
For instance, if the true label distribution of $\mathcal{D}_k$ is $[0.1, 0.1, 0.8]$, the estimated distribution may instead be $[0.6, 0.05, 0.35]$, yielding low accuracy and failing to reflect the ground-truth distribution. 
To address this, we propose a cooperative data selection strategy that redistributes adjusted pseudo-label distributions to clients in proportion to their locally estimated distributions.

To obtain representative pseudo-label distributions, each client computes category-wise statistics on its filtered pseudo-labels, yielding an estimated distribution $U^\text{est}_k \in \mathbb{R}^C$, where the $c$-th element $u_{k,c} = \sum_{i} \mathbbm{1}(\hat{y}_i = c)$ counts the number of samples assigned to class $c$ in client $k$. 
Clients then upload $U^\text{est}_k$ to the server for collaborative allocation. 
Following prior works~\cite{ye2023feddisco,zhang2024candidate}\nocite{yu2026fedbst}, we assume a uniform joint label distribution across all clients. 
To ensure balanced training across all categories, the server globally selects $M$ pseudo-labels per class and allocates them to clients based on their estimated distributions:
\begin{align}
\widetilde{U}_k &= (\widetilde{u}_{k,1}, \ldots, \widetilde{u}_{k,C}), \\
\widetilde{u}_{k,c} &= \left\lceil \frac{u_{k,c}}{\sum_{i=1}^{K} u_{i,c}} \cdot M \right\rceil,
\label{eq:count}
\end{align} 
where $M=\frac{1}{C} \sum_{k=1}^K\sum_{c=1}^C {u_{k,c}}$ is the number of pseudo-labels allocated to each class globally, and $\widetilde{u}_{k,c}$ denotes the number of pseudo-labels of class $c$ assigned to client $k$.
Each client then constructs its training set $\widetilde{D}_k$ based on the allocated counts $\widetilde{U}_{k}$.

Specifically, for each class $c$, the $\widetilde{u}_{k,c}$ samples with the highest prediction probabilities are selected from $\mathcal{D}_k$ and added to $\widetilde{\mathcal{D}}_k$. 
As shown in Eq.~(\ref{eq:count}), the total budget $M$ is allocated proportionally based on each client's estimated distribution per category, rather than its absolute values. 
This strategy ensures that even when the estimated distributions are noisy or fail to fully reflect the client's true label distribution, the selected pseudo-labels remain representative and accurate, preserving the essential characteristics of the client's data and supporting robust collaborative learning.

As federated training progresses, we periodically repeat the complete cycle of pseudo-label generation, distribution estimation, global allocation, and local data selection.
As shown in Fig.~\ref{fig:framework}, this iterative process keeps $\widetilde{\mathcal{D}}_k$ regularly updated, ensuring that the pseudo-labeled dataset remains globally balanced and of high quality.

\subsection{Dual Pseudo-labeling}
\label{sec:local}
Having obtained reliable samples through cooperative data selection, we now address the problem of fine-tuning the model with these selected data on each client.
A natural strategy is to generate pseudo-labels for the sampled instances and optimize the prompts using a cross-entropy objective. 
However, the allocated pseudo-labels may still exhibit class imbalance. 
To address this, we first compute the global frequency of class $c \in [1, \dots, C]$ at the server as follows:
\begin{equation}
d_c = \frac{\sum_{k=1}^K \widetilde{u}_{k,c}}{\sum_{k=1}^K \sum_{i=1}^C \widetilde{u}_{k,i}}.
\end{equation}
We then define inverse-frequency weights to upweight rare classes, while clipping extreme values to prevent gradient explosion:
\begin{equation}
w_c = \min \left( \frac{1}{d_c \cdot C}, w_{\max} \right),
\end{equation}
with $w_{\max}=3.0$ by default. 
The final weighted pseudo-labeling loss is formulated as:
\begin{equation}
\mathcal{L}_\texttt{pos} = \mathbb{E}_{(x,\hat{y}) \in \widetilde{\mathcal{D}}_k} \left[ \ell_{ce}(x, \hat{y}) \cdot w_{\hat{y}} \right],
\end{equation}
where $w_{\hat{y}}$ denotes the weight associated with the pseudo-label class $\hat{y}$.

To maximize the utility of model predictions, we draw on prior works~\cite{rizve2021defense,wang2022semi,wang2024curriculum} that constrain the probabilities of less likely classes for unlabeled data. 
These low-probability predictions are known as negative pseudo-labels, in contrast to the high-confidence positive pseudo-labels. 
Motivated by this line of research, we incorporate a negative pseudo-labeling loss into our framework. Specifically, for each sample $x$, we define $\mathcal{Y}_{neg}(x)$ as the set of the $B$ least probable classes:
\begin{equation}
\mathcal{Y}_\texttt{neg}(x) = \arg\min_{|\mathcal{N}| = B} \sum_{c \in \mathcal{N}} p_c(x) 
\end{equation}
where $B = \lceil C \times \tau_{\text{neg}} \rceil$ and $\tau_{\text{neg}} \in (0,1)$ controls the proportion of classes selected as negative. The negative loss for client $k$ is then defined as:
\begin{equation}
\mathcal{L}_\texttt{neg} = - \mathbb{E}_{x \in \widetilde{\mathcal{D}}_k} \sum_{y \in \mathcal{Y}_\texttt{neg}(x)} \log \left( 1 - p_y(x) \right).
\end{equation}
This loss encourages the model to push down probabilities on unlikely classes, thereby improving the clarity and discriminability of the predictions.

Finally, we formulate the dual pseudo-labeling loss for local training at each client as the combination of the positive and negative pseudo-labeling losses:
\begin{equation}
\mathcal{L}_\texttt{total} = \mathcal{L}_\texttt{pos} + \mathcal{L}_\texttt{neg}.
\end{equation}

\begin{figure}
\small
\setlength\tabcolsep{2mm}
\begin{tabular}{cc}
\includegraphics[width=0.445\linewidth]{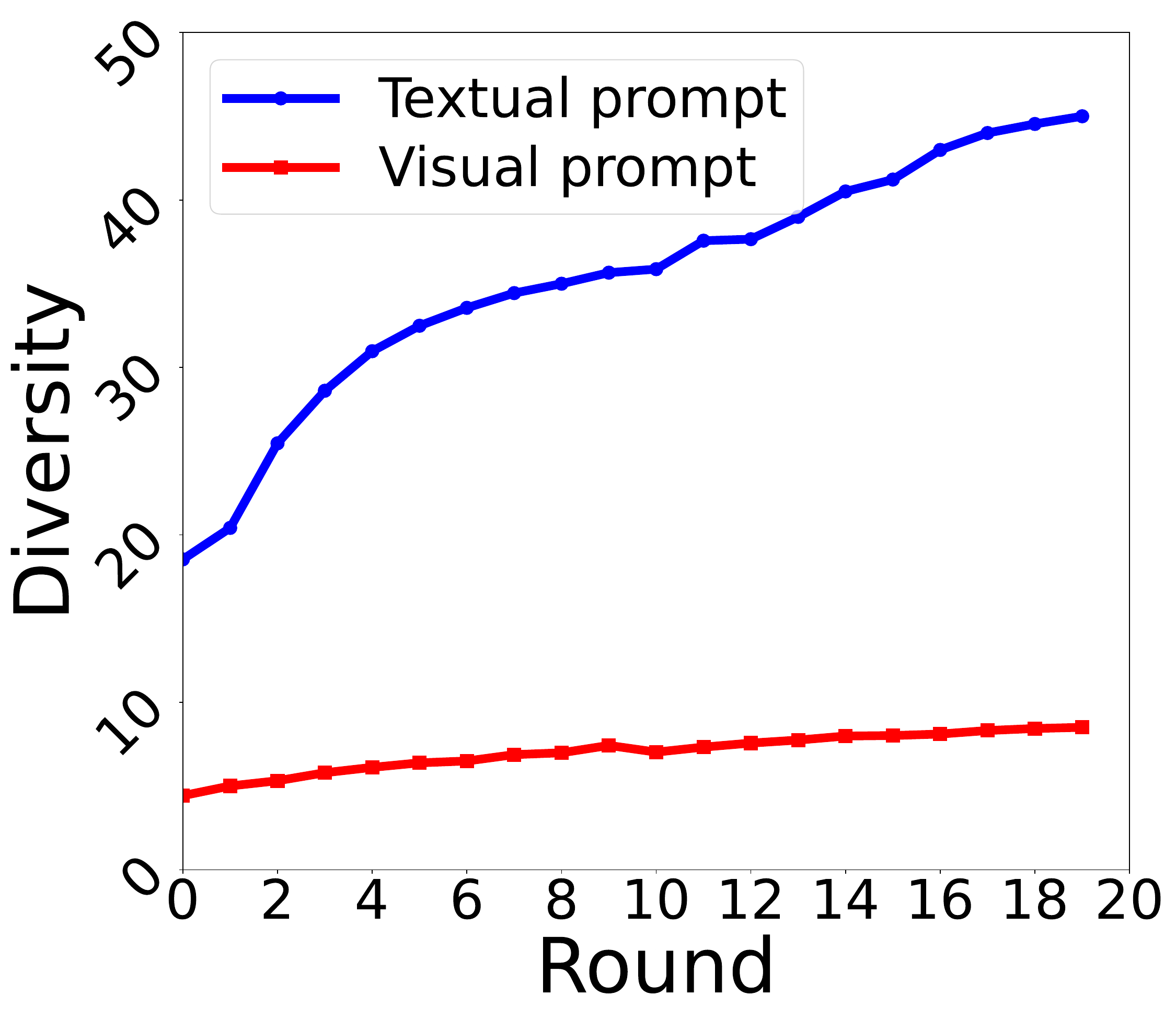} &		
\includegraphics[width=0.445\linewidth, clip]{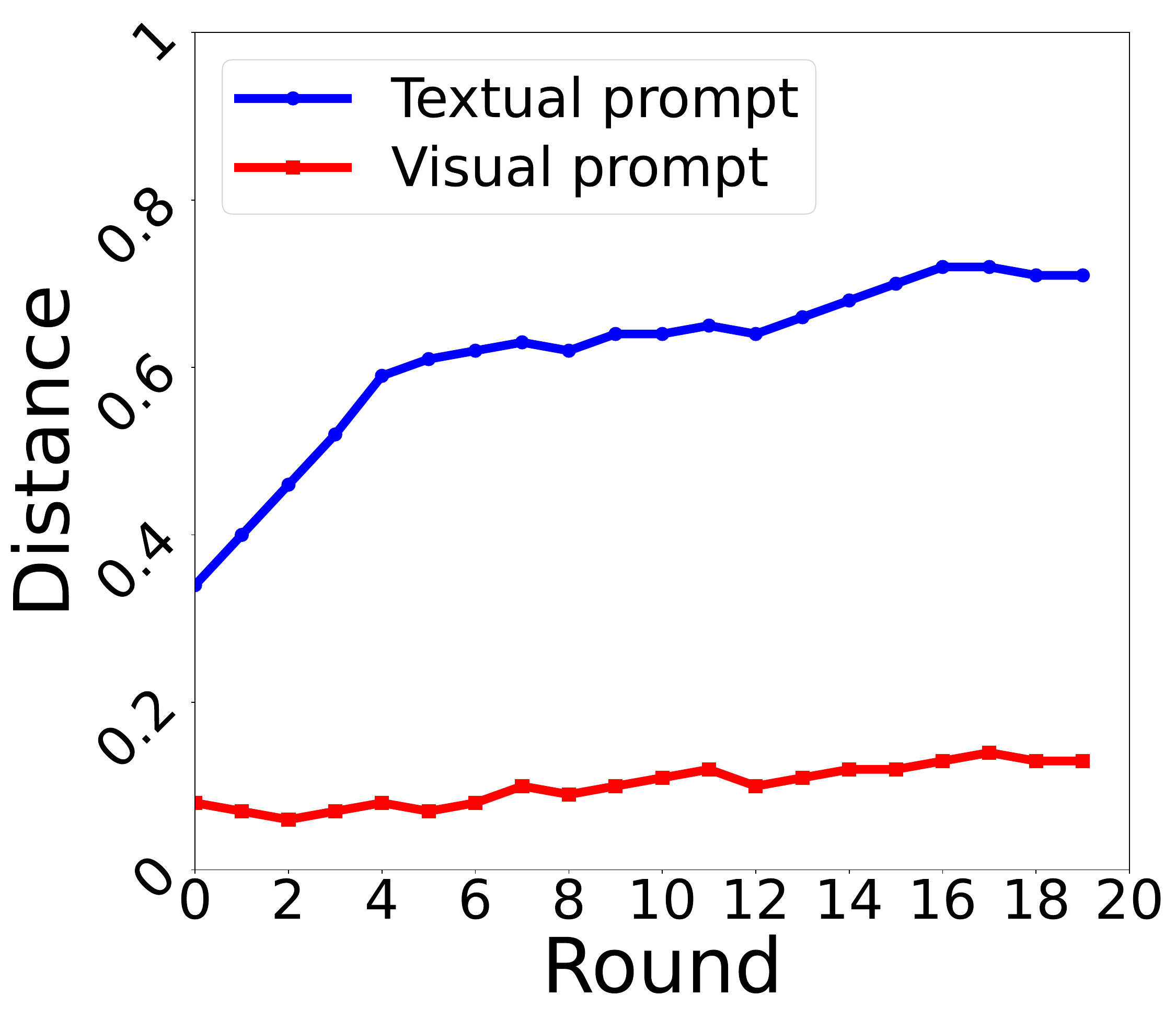} 
\\
(a) Drift Diversity & (b) Cosine Distance
\end{tabular}
\caption{Drift diversity and cosine distance of prompts among clients during training on CIFAR-10~\cite{krizhevsky2009learning}. The observed differences in textual prompts are notably greater than those in visual prompts.}
\label{fig:shift}
\end{figure}

\subsection{Personalized Prompt Aggregation}
\label{sec:prompt}
For each client, we optimize both the textual prompt $P^t_k$ and the visual prompt $P^v_k$ using the local training set $\widetilde{\mathcal{D}}_k$.
Since all clients share the same task, aggregating their learned prompts offers a natural avenue for performance improvement.
A simple averaging operation across clients appears to be a straightforward solution.
However, due to data heterogeneity, especially label skew across clients, this may lead to suboptimal or even detrimental performance.

In CLIP, updates to the visual branch primarily enhance image representations, while the textual branch focuses on determining classification boundaries using category information~\cite{xing2023dual}.
Under heterogeneous data distributions across clients, visual prompts $\{P^v_k\}_{k=1}^K$ are more likely to be similar, whereas textual prompts $\{P^t_k\}_{k=1}^K$ exhibit greater variability.
To validate this conjecture, we measure the differences for each prompt across all clients using drift diversity~\cite{li2023effectiveness} and cosine distance, which capture the magnitude and direction of prompt updates, respectively.
As shown in Fig.~\ref{fig:shift}, textual prompts exhibit significantly greater divergence than visual prompts, confirming our hypothesis.

To avoid conflicts in aggregating widely divergent textual prompts while promoting collaboration, we propose a personalized prompt aggregation protocol.
In this protocol, all clients upload their visual prompts $\{P^v_k\}_{k=1}^K$ to the server for aggregation, while retaining their textual prompts $\{P^t_k\}_{k=1}^K$ locally for personalization.
On the server side, we aggregate the visual prompts using a weighted average following~\cite{mcmahan2017communication}:
\begin{equation}
P^v = \sum_{k=1}^K \frac{\widetilde{n}_k}{\sum_{i=1}^K \widetilde{n}_i} \cdot P^v_k,
\end{equation}
where $\widetilde{n}_k = \sum_c \widetilde{u}_{k,c}$ denotes the total number of pseudo-labeled samples assigned to client $k$.
The aggregated visual prompts $P^v$ are then distributed back to all clients to initialize their local models for the next training round, as shown in Fig.~\ref{fig:framework}.

\textbf{Remark}. 
FedUR preserves client privacy by ensuring that the uploaded estimated distributions are derived solely from CLIP's predictions, making it infeasible to inversely infer exact label distributions. 
This highlights the enhanced privacy-preserving nature of our method, particularly compared to existing approaches~\cite{zhang2022fine,kim2022federated} that require transmitting exact category distributions and thus risk exposing sensitive client information.
Moreover, unlike methods that aggregate prompts from both modalities, our approach reduces communication overhead by uploading only visual prompts, making it more practical for bandwidth-constrained environments.

\begin{algorithm}
\renewcommand{\algorithmicrequire}{\textbf{Input:}}
\renewcommand{\algorithmicensure}{\textbf{Output:}}
\caption{\textbf{FedUR}}
\label{alg:fedur}
\begin{algorithmic}[1]
\Require number of communication rounds $T$, number of clients $K$, unlabeled dataset $\{\mathcal{D}_k\}^K_{k=1}$, client participating rate $R$, number of local epochs $E$, batch size $B$, learning rate $\eta$, pseudo labels update interval $Q$.
\Ensure global visual prompt $P^v$ and personalized textual prompts $\{P^t_k\}^K_{k=1}$
\State initialize $P^{v,0}$ , $\{P^{t,0}_k\}^K_{k=1}$
\State $m \gets \max( \lfloor R \cdot K \rfloor, 1)$
\For{\textit{communication round} $r=1,2,\cdots,T$}
\If{r \% Q = 0} 
\State {\{$\{\widetilde{\mathcal{D}}_k\}, d$\} $\gets$ \textbf{data\_selection}($\{\mathcal{D}_k\}^K_{k=1}$)}
\EndIf
\State $M_r \gets $ randomly select $m$ clients 
\For{\textit{each client} $k \in M_r$}
\State $P^{v,r}_k = P^{v,r}$
\State $P^{v,r+1}_k, P^{t,r+1}_k \gets$ \textbf{dual\_label}($P^{v,r}_k, P^{t,r}_k, d$)
\EndFor
\State $P^{v,r+1} = P^{v,r} + \sum_{k \in M_r}{\frac{|\widetilde{\mathcal{D}}_k|}{|\widetilde{\mathcal{D}}|}(P^{v,r+1}_k - P^{v,r}_k})$
\EndFor 
\vspace{0.5cm}

\State \textbf{data\_selection}($\{\mathcal{D}_k\}^K_{k=1}$):
\State {\color{gray}{\# cooperative data selection in Sec.~\ref{sec:coper}}}
\For {$k=1, ..., K$}
\State {$\mathcal{D}^{est}_k = \{(x, \hat{y}) |\max\limits_{c} p_c(x) > \tau_1 , \texttt{Ent}(p_c(x)) < \tau_2\}$}, 
\State $\widetilde{U}_k = \{\widetilde{u}_{k,1}, \widetilde{u}_{k,2}, ..., \widetilde{u}_{k,C}\}, \widetilde{u}_{k,c} = \frac{u_{k,c}}{\sum^{K}_{k=1}{u_{k,c}}} \times M$, $u_{k,c}$ is the number of data for class $c$ in $\mathcal{D}^{est}_k$,
\State $\widetilde{\mathcal{D}}_k$ is selected by the most confident samples according to the capacity $\widetilde{U}_{k}$
\State $d=\{d_1, d_2, ..., d_C\}$ is the global pseudo label distribution, where $d_c = \frac{\sum_{k=1}^K \widetilde{u}_{k,c}}{\sum_{k=1}^K \sum_{i=1}^C \widetilde{u}_{k,i}}$
\EndFor
\State \textbf{return} $\{\{\widetilde{\mathcal{D}}_k\}^K_{k=1}, d\}$

\vspace{0.5cm}
\State \textbf{dual\_label}($P^{v,r}_k, P^{t,r}_k, d$):
\State {\color{gray}{\# dual pseudo-labeling in Sec.~\ref{sec:local}}}
\State
$w_c = \min \left( \frac{1}{d_c \cdot C}, w_{\max} \right)$
\For{\textit{epoch} $e=1,2,\cdots,E$}
 \For{\textit{each batch} $\mathcal{B}_{i} = \{\textit{x},\hat{y}\} \in \widetilde{\mathcal{D}}_i$} 
 \State $\mathcal{L}_\texttt{pos} = \mathbb{E}_{(x,\hat{y}) \in \widetilde{\mathcal{D}}_k} \left[ \ell_{ce}(x, \hat{y}) \cdot w_{\hat{y}} \right]$,
 \State $\mathcal{Y}_\texttt{neg}(x) = \arg\min_{|\mathcal{N}| = B} \sum_{c \in \mathcal{N}} p_c(x) $,
 \State $\mathcal{L}_\texttt{neg} = - \mathbb{E}_{x \in \widetilde{\mathcal{D}}_k} \sum_{y \in \mathcal{Y}_\texttt{neg}(x)} \log \left( 1 - p_y(x) \right)$.
 \State $\mathcal{L}_\texttt{total} = \mathcal{L}_\texttt{pos} + \mathcal{L}_\texttt{neg}.$
 \vspace{0.4\baselineskip}
 \State $P^{v,r}_k = P^{v,r}_k - \eta \nabla {\mathcal{L}_\texttt{total}}$
 \State $P^{t,r}_k = P^{t,r}_k - \eta \nabla {\mathcal{L}_\texttt{total}}$
 \EndFor
\EndFor 
\State \textbf{return} $P^{v,r}_k, P^{t,r}_k$
\end{algorithmic}
\end{algorithm}

\section{Experiments}
\label{sec:experiment}

\subsection{Setups}
\textbf{Datasets.} We evaluate our method on eight public datasets spanning three types of data heterogeneity: label skew, feature skew, and their combination. 
For label skew, we follow prior works~\cite{li2024global,guo2025cooperative} and employ four visual classification datasets—DTD~\cite{cimpoi2014describing}, RESISC45~\cite{cheng2017remote}, UCF101~\cite{soomro2012ucf101}, and CUB~\cite{wah2011caltech}, along with CIFAR10 and CIFAR100~\cite{krizhevsky2009learning} as standard federated benchmarks. 
Two label skew scenarios are simulated: Dirichlet-based and quantity-based. 
In Dirichlet-based skew, each client's class samples are drawn from a Dirichlet distribution parameterized by $\beta$, where smaller $\beta$ indicates more severe skew. 
In quantity-based skew, training data are grouped by label and partitioned into shards with imbalanced sizes, with each client receiving a portion and each shard covering 20\% of categories by default.
For feature skew, we use DomainNet~\cite{peng2019moment} (6 domains: Clipart ({C}), Infograph ({I}), Painting ({P}), Quickdraw ({Q}), Real ({R}), and Sketch ({S})) and OfficeHome~\cite{venkateswara2017deep} (4 domains: Art ({A}), Clipart ({C}), Product ({P}), and Real\_World ({R})), assigning data from each domain to a distinct client. 
To evaluate combined label and feature skew, we distribute samples of each domain across 3 and 5 clients using Dirichlet distributions with $\beta \in \{0.1, 0.05\}$.

\begin{table*}[t]
\setlength{\tabcolsep}{2pt}
\centering
\caption{Performance (\%) under \textbf{Dirichlet-based label skews} with ViT/B-32 as the backbone. 
FPL~\cite{menghini2023enhancing}, CPL~\cite{zhang2024candidate}, and CoPL~\cite{guo2025cooperative} are adopted as pseudo-labeling (\textbf{PL}) methods. \textbf{Bold} and \underline{underline} represent the best and second-best results.}
\label{tab:accuracy-dirichlet}
\resizebox{\textwidth}{!}{ 
\begin{tabular}{lccccccccccccccc}
\toprule 
\multirow{2}{*}{Method} & \multirow{2}{*}{\textbf{PL}} & \multicolumn{2}{c}{{DTD}} & \multicolumn{2}{c}{{RESISC45}} & \multicolumn{2}{c}{{CUB}} & \multicolumn{2}{c}{{UCF101}} & \multicolumn{2}{c}{{CIFAR10}} & \multicolumn{2}{c}{{CIFAR100}} & \multicolumn{2}{c}{{Average}} \\ \cmidrule{3-16} 
& & $\beta = 0.1$ & $\beta = 0.05$ & $\beta = 0.1$ & $\beta = 0.05$& $\beta = 0.1$ & $\beta = 0.05$& $\beta = 0.1$ & $\beta = 0.05$& $\beta = 0.1$ & $\beta = 0.05$ & $\beta = 0.1$ & $\beta = 0.05$ & $\beta = 0.1$ & $\beta = 0.05 $\\ \midrule
CLIP & - & {43.24} &{43.24} & {54.51} & {54.51} & 51.28 & 51.28 & {61.00} & {61.00} & {86.93} & {86.93} & {64.17} & {64.17} & 60.19 & 60.19 \\ 	
PromptFL~\cite{guo2023promptfl} & FPL & 45.79 & 44.62 & 59.76 & 58.05 & 47.29 & 46.16 & 64.39 & 62.96 & 87.51 & 85.02 & 63.26 & 62.56 & 61.33 & 59.90 \\	
PromptFL~\cite{guo2023promptfl} & CPL & 44.84 & 46.32 & 62.52 & 60.99 & 48.72 & 48.86 & 63.86 & 64.97 & 88.90 & 89.14 & 66.04 & 65.14 & 62.48 & 62.57 \\
Promptprox~\cite{li2020federated} & FPL & 45.15 & 43.88 & 59.36 & 58.59 & 47.04 & 47.01 & 62.91 & 61.27 & 87.13 & 85.62 & 62.77 & 64.12 & 60.73 & 60.08 \\
Promptprox~\cite{li2020federated} & CPL & 43.51 & 45.85 & 59.99 & 62.35 & 49.25 & 48.65 & 64.55 & 63.94 & 90.44 & 90.07 & 65.86 & 66.15 & 62.27 & 62.84 \\	
pFedPrompt~\cite{guo2023pfedprompt} & FPL & 44.56 & 46.19 & 65.95 & 60.52 & 48.48 & 44.42 & 64.37 & 64.18 & 87.45 & 90.72 & 65.08 & 65.83 & 62.65 & 61.98 \\	
pFedPrompt~\cite{guo2023pfedprompt} & CPL & 44.22 & 47.59 & 61.76 & 66.79 & 47.23 & 50.97 & 65.59 & 65.44 & 90.26 & 89.10 & 65.63 & 67.86 & 62.45 & 64.63 \\	
FedOPT~\cite{li2024global} & FPL & 51.97 & 48.71 & 57.76 & 55.62 & 47.99 & 47.05 & 61.83 & 64.56 & 91.32 & 90.73 & 67.51 & 66.50 & 63.06 & 62.20 \\	
FedOPT~\cite{li2024global} & CPL & 37.43 & 39.33 & 51.10 & 48.47 & 46.61 & 45.07 & 58.93 & 57.12 & 91.08 & 91.36 & 57.87 & 59.81 & 57.17 & 56.86 \\ 
FedCoPL~\cite{guo2025cooperative} & CoPL & \underline{60.89} & \underline{66.37} & \underline{75.76} & \underline{80.26} & \underline{56.09} & \underline{54.80} & \underline{73.20} & \underline{74.97} & \underline{95.38} & \underline{95.11} & \underline{73.59} & \underline{72.84} & \underline{72.49} & \underline{74.06} \\ \cmidrule{1-16} \rowcolor[gray]{0.9} 
\textbf{FedUR} & CoDPL & \textbf{62.75} & \textbf{69.30} & \textbf{78.61} & \textbf{82.01} & \textbf{57.63} & \textbf{55.09} & \textbf{74.29} & \textbf{76.75} & \textbf{96.41} & \textbf{96.06} & \textbf{73.81} & \textbf{73.54} & \textbf{73.92} & \textbf{75.46} \\
\bottomrule
\end{tabular}
}
\end{table*}

\begin{table*}[t]
\setlength{\tabcolsep}{8pt}
\centering
\caption{Performance (\%) under \textbf{quantity-based label skews} with ViT/B-32 as the backbone.
}
\label{tab:accuracy-quantity}
\resizebox{0.8\textwidth}{!}{ 
\begin{tabular}{lcccccccc}
\toprule
{Method} & \textbf{PL} & {{DTD}} & {{RESISC45}} & {{CUB}} & {{UCF101}} & {{CIFAR10}} & {{CIFAR100}} & {Average} \\ \midrule
CLIP & - & {43.24} & {54.51} & 51.28 & {61.00} & {86.93} & {64.17} & 60.19 \\ 
PromptFL~\cite{guo2023promptfl} & FPL & 43.53 & 57.53 & 46.66 & 62.35 & 88.29 & 63.94 & 60.38 \\ 
PromptFL~\cite{guo2023promptfl} & CPL & 43.82 & 61.11 & 47.31 & 64.23 & 88.70 & 67.23 & 62.07 \\
Promptprox~\cite{li2020federated} & FPL & 44.89 & 56.71 & 48.74 & 61.80 & 85.69 & 64.01 & 60.31 \\
Promptprox~\cite{li2020federated} & CPL & 45.53 & 60.77 & 47.61 & 63.86 & 89.21 & 66.49 & 62.25 \\
pFedPrompt~\cite{guo2023pfedprompt} & FPL & 45.09 & 60.46 & 47.71 & 65.39 & 87.54 & 62.82 & 61.50\\
pFedPrompt~\cite{guo2023pfedprompt} & CPL & 45.14 & 66.68 & 50.45 & 65.57 & 87.98 & 65.45 & 63.55 \\
FedOPT~\cite{li2024global} & FPL & 47.65 & 56.04 & 46.61 & 62.95 & 90.74 & 66.76 & 61.79 \\
FedOPT~\cite{li2024global} & CPL & {40.24} & 47.09 & 44.00 & 57.24 & 89.21 & 57.98 & 55.96 \\
FedCoPL~\cite{guo2025cooperative} & CoPL & \underline{56.18} & \underline{81.06} & \underline{56.31} & \underline{72.03} & \underline{94.86} & \underline{73.39} & \underline{72.31} \\ \cmidrule{1-9} \rowcolor[gray]{0.9}
\textbf{FedUR} & CoDPL & \textbf{58.60} & \textbf{82.56} & \textbf{57.66} & \textbf{72.56} & \textbf{95.03} & \textbf{73.60} & \textbf{73.34} \\
\bottomrule 	 	 	 	 		 
\end{tabular}
}
\end{table*}

\textbf{Baseline methods.}
We compare our method (FedUR) against a comprehensive set of baselines, including FedCoPL~\cite{guo2025cooperative} and the methods reported therein. 
For pseudo-label selection strategies, FPL~\cite{menghini2023enhancing} selects the most reliable samples per class based on prediction confidence, while CPL~\cite{zhang2024candidate} assigns multiple candidate pseudo-labels via joint sample- and category-level selection.
For supervised federated learning, PromptFL~\cite{guo2023promptfl} learns shared prompts via FedAvg~\cite{mcmahan2017communication}; PromptProx~\cite{guo2023promptfl}, adapted from FedProx~\cite{li2020federated}, incorporates a proximal regularization term; pFedPrompt~\cite{guo2023pfedprompt} augments a unified textual prompt with personalized attention modules; and FedOPT~\cite{li2024global} performs conditional optimal transport between global and local prompts. 
FedCoPL~\cite{guo2025cooperative} is included as a strong unsupervised federated baseline.
We denote the cooperative data selection strategy, combined with dual pseudo-labeling, as CoDPL in our experiments.
As a reference, we also report zero-shot CLIP performance, where a pre-trained CLIP model with a hand-crafted prompt template (\eg, ``\texttt{a photo of a [CLASS]}") is directly applied at test time.

\textbf{Implementation details.}
All experiments are conducted with a frozen CLIP model. 
We set the number of clients $K$ to 5 for CUB~\cite{wah2011caltech} and UCF101~\cite{soomro2012ucf101}, and to 10 for all other datasets, with full client participation by default. 
For DomainNet and OfficeHome, we distribute samples from each domain across 1, 3, and 5 clients. 
We perform 20 communication rounds, with pseudo-labels updated every 5 rounds using the latest local model. 
Within each round, clients train locally for 10 epochs using mini-batch stochastic gradient descent (SGD) with a learning rate of 0.1, momentum of 0.9, and cosine annealing decay. 
We report mean accuracy averaged over three independent trials.
Unless otherwise specified, ViT-B/32 is adopted as the backbone, with $\tau_{neg}=0.2$ by default.
All experiments are implemented in PyTorch~\cite{paszke2019pytorch} on NVIDIA 3090 GPUs.

\begin{table*}[t]
\setlength{\tabcolsep}{8pt}
\centering
\caption{Performance (\%) under \textbf{feature skew} with ViT/B-32 as the backbone on DomainNet and OfficeHome datasets.
}
\label{tab:accuracy-feature}
\resizebox{\textwidth}{!}{ 
\begin{tabular}{lccccccccccccc}
\toprule 
\multirow{2}{*}{Method} & \multirow{2}{*}{\textbf{PL}} & \multicolumn{7}{c}{DomainNet} & \multicolumn{5}{c}{OfficeHome}\\
\cmidrule{3-9} \cmidrule{10-14}
& & C & I & P & Q & R & S & Average & A & C & P & R & Average \\ \midrule
CLIP & - & 97.72 & 69.56 & \underline{96.28} & 66.00 & 96.63 & 95.31 & 86.92 & 77.32 & 64.84 & 88.41 & 90.03 & 80.15 \\ 
PromptFL~\cite{guo2023promptfl} & FPL & 97.34 & 73.06 & 93.86 & 67.00 & 96.63 & \underline{95.85} & 87.29 & 79.03 & 69.75 & 88.70 & 90.72 & 82.05 \\ 
PromptFL~\cite{guo2023promptfl} & CPL & 96.96 & 72.91 & 95.64 & \textbf{73.50} & \textbf{97.12} & 94.58 & \underline{88.45} & 78.91 & 69.52 & 89.15 & 90.84 & 82.11 \\
Promptprox~\cite{li2020federated} & FPL & \textbf{97.91} & \textbf{74.89} & \textbf{96.45} & 67.00 & \underline{96.96} & 95.31 & 88.09 & 78.38 & 69.23 & 89.21 & 90.61 & 81.86 \\
Promptprox~\cite{li2020federated} & CPL & 97.15 & 73.97 & 94.67 & 71.10 & 96.71 & 94.95 & 88.09 & 77.32 & 66.44 & 88.08 & 87.97 & 79.95 \\
pFedPrompt~\cite{guo2023pfedprompt} & FPL & 89.92 & 66.06 & 88.53 & 63.50 & 93.02 & 88.27 & 81.55 & \underline{79.62} & \underline{69.98} & 88.43 & 90.07 & 82.03 \\
pFedPrompt~\cite{guo2023pfedprompt} & CPL & 96.58 & 72.30 & 95.96 & 67.10 & 96.38 & 93.68 & 87.00 & 78.11 & \textbf{70.09} & \underline{89.43} & 90.38 & 82.00 \\
FedOPT~\cite{li2024global} & FPL & 97.34 & \underline{74.73} & 96.12 & 63.70 & 96.80 & \textbf{96.03} & 87.45 & 78.38 & 69.66 & 89.21 & 90.84 & 82.02 \\
FedOPT~\cite{li2024global} & CPL & 97.15 & 70.62 & 94.51 & 71.40 & 96.63 & 94.22 & 87.42 & 78.59 & 69.29 & 88.54 & \underline{90.85} & 81.82 \\
FedCoPL~\cite{guo2025cooperative} & CoPL & 97.15 & 72.60 & 95.64 & 70.60 & 96.38 & 94.58 & 87.83 & \textbf{80.03} & 69.55 & 89.31 & 89.92 & \underline{82.20}\\ \midrule
\rowcolor[gray]{0.9} \textbf{FedUR} & CoDPL & \underline{97.84} & 73.06 & \underline{96.28} & \underline{72.10} & 96.47 & 95.77 & \textbf{88.59} & 79.18 & 69.41 & \textbf{89.88} &\textbf{90.86} & \textbf{82.33} \\
\bottomrule 	 	 	 	 		 
\end{tabular}
}
\end{table*}

\begin{table*}[t]
\setlength{\tabcolsep}{2pt}
\centering
\caption{Performance (\%) under \textbf{domain and label skews} with ViT/B-32 as the backbone on DomainNet dataset. For each domain, the dataset is split into three clients according to a Dirichlet distribution ($\beta$ = 0.05 or 0.1).}
\label{tab:accuracy-label-feature-domain}
\resizebox{\textwidth}{!}{ 
\begin{tabular}{lccccccccccccccc}
\toprule 
\multirow{2}{*}{Method} & \multirow{2}{*}{\textbf{PL}} & \multicolumn{2}{c}{{C}} & \multicolumn{2}{c}{{I}} & \multicolumn{2}{c}{{P}} & \multicolumn{2}{c}{{Q}} & \multicolumn{2}{c}{{R}} & \multicolumn{2}{c}{{S}} & \multicolumn{2}{c}{{Average}} \\ \cmidrule{3-16} 
& & $\beta = 0.1$ & $\beta = 0.05$ & $\beta = 0.1$ & $\beta = 0.05$& $\beta = 0.1$ & $\beta = 0.05$& $\beta = 0.1$ & $\beta = 0.05$& $\beta = 0.1$ & $\beta = 0.05$ & $\beta = 0.1$ & $\beta = 0.05$ & $\beta = 0.1$ & $\beta = 0.05 $\\ \midrule
CLIP & - & 97.72 & 97.72 & 69.56 & 69.56 & 96.28 & \underline{96.28} & 66.00 & 66.00 & 96.63 & 96.63 & 95.31 & 95.31 & 86.92 & 86.92 \\ 
PromptFL~\cite{guo2023promptfl} & FPL & 96.72 & 72.10 & 73.21 & 54.92 & \textbf{97.08} & 72.08 & 60.91 & 52.49 & 96.83 & 78.85 & 93.46 & 67.42 & 86.37 & 66.31 \\ 
PromptFL~\cite{guo2023promptfl} & CPL & 95.09 & 94.43 & 69.68 & 66.49 & 92.91 & 90.77 & 69.06 & 67.95 & 95.55 & 93.34 & 91.50 & 90.60 & 85.63 & 83.93 \\
Promptprox~\cite{li2020federated} & FPL & 84.55 & 68.99 & \underline{74.29} & 53.25 & 86.12 & 73.98 & 53.78 & 46.79 & 84.86 & 78.12 & 85.90 & 67.72 & 78.25 & 64.81 \\
Promptprox~\cite{li2020federated} & CPL & 90.69 & 93.55 & 64.41 & 66.83 & 91.56 & 92.54 & 59.55 & 60.22 & 94.98 & 93.90 & 90.45 & 91.94 & 81.94 & 83.16 \\
pFedPrompt~\cite{guo2023pfedprompt} & FPL & \textbf{98.73} & \textbf{98.71} & 73.38 & \textbf{76.75} & 96.13 & \textbf{96.99} & 66.19 & 66.12 & 96.90 & 97.15 & 96.80 & \underline{96.44} & 88.02 & 88.69 \\
pFedPrompt~\cite{guo2023pfedprompt} & CPL & 87.03 & 81.93 & 48.20 & 64.44 & 81.36 & 86.56 & 53.72 & 56.54 & 90.79 & 87.08 & 79.97 & 80.36 & 73.51 & 76.15 \\
FedOPT~\cite{li2024global} & FPL & \underline{98.48} & 95.95 & \textbf{74.30} & 69.89 & \underline{97.07} & 95.36 & 69.09 & 49.12 & 97.30 & 95.39 & \textbf{97.22} & 94.11 & \textbf{88.91} & 83.30 \\
FedOPT~\cite{li2024global} & CPL & 80.31 & 76.82 & 52.70 & 50.74 & 75.23 & 71.75 & 45.45 & 44.96 & 75.69 & 77.48 & 73.22 & 67.46 & 67.10 & 64.87 \\
FedCoPL~\cite{guo2025cooperative} & CoPL & 97.64 & 97.79 & 72.22 & 73.34 & 95.56 & 95.59 & \textbf{72.52} & \underline{72.41} & \textbf{97.54} & \underline{97.47} & \underline{96.92} & 96.02 & 88.73 & \underline{88.77} \\ \cmidrule{1-16} \rowcolor[gray]{0.9} 
\textbf{FedUR} & CoDPL & 97.80 & \underline{98.31} & 72.82 & \underline{73.63} & 95.87 & 95.55 & \underline{71.43} & \textbf{72.47} & \underline{97.32} & \textbf{97.96} & \textbf{97.22} & \textbf{96.61} & \underline{88.75} & \textbf{89.09} \\
\bottomrule 
\end{tabular}
}
\end{table*}

\subsection{Experimental Results}
\textbf{Results under label skews.}
Table~\ref{tab:accuracy-dirichlet} and Table~\ref{tab:accuracy-quantity} present the performance of all methods under Dirichlet-based and quantity-based label skews, respectively.
For Dirichlet-based skew ($\beta \in \{0.1, 0.05\}$), FedUR consistently outperforms all baselines across every dataset, demonstrating strong robustness to probabilistic label imbalance. 
Among the baselines, FedCoPL achieves substantially better performance than other methods through its cooperative pseudo-labeling and partial prompt aggregation, confirming its effectiveness in mitigating heterogeneity. 
FedUR further improves upon FedCoPL across all datasets and skew levels, indicating that the incorporation of negative pseudo-labels and refined pseudo-label distribution yields additional benefits.
For quantity-based skew (shard size $s=C\times 0.2$, where $C$ is the number of classes), each client only contains samples from a limited subset of classes, posing greater challenges for pseudo-label generation and optimization. 
Under this more severe setting, several baselines fail to outperform zero-shot CLIP, highlighting their sensitivity to class sparsity. 
FedCoPL remains a strong unsupervised baseline, while FedUR consistently achieves the best performance across all datasets. 
These results validate that our approach is robust to both probabilistic and extreme label skews, confirming its effectiveness and general applicability in heterogeneous federated learning.

\textbf{Results under feature skew.} 
Table~\ref{tab:accuracy-feature} reports the performance of all methods under feature skew on the DomainNet and Office-Home datasets, respectively, using ViT-B/32 as the backbone.
Feature skew introduces significant distribution discrepancies across clients, posing substantial challenges for federated learning with frozen image encoders. 
Under this setting, several baseline methods perform comparably to or worse than zero-shot CLIP, which can be attributed to their inability to effectively bridge the gap between heterogeneous visual features and textual prompts. 
Among the baselines, FedCoPL achieves competitive performance through cooperative pseudo-labeling, demonstrating its effectiveness in alleviating heterogeneity to some extent. 
Our method, FedUR, consistently achieves the best performance across both datasets, outperforming all competing approaches by a clear margin. 
Notably, the gains are consistent across multiple domains in DomainNet and all domains in Office-Home, highlighting the robustness of FedUR to severe feature skew. 
These results confirm the effectiveness and general applicability of FedUR for handling feature-skew scenarios.

\begin{table*}[t]
\setlength{\tabcolsep}{4pt}
\centering
\caption{Performance (\%) under \textbf{domain and label skews} with ViT/B-32 as the backbone on DomainNet dataset. For each domain, the dataset is split into three clients according to a Dirichlet distribution ($\beta$ = 0.05 or 0.1).}
\label{tab:accuracy-label-feature-office}
\resizebox{\textwidth}{!}{ 
\begin{tabular}{lcccccccccccc}
\toprule 
\multirow{2}{*}{Method} & \multirow{2}{*}{\textbf{PL}} & \multicolumn{2}{c}{{A}} & \multicolumn{2}{c}{{C}} & \multicolumn{2}{c}{{P}} & \multicolumn{2}{c}{{R}} & \multicolumn{2}{c}{{Average}} \\ \cmidrule{3-12} 
& & $\beta = 0.1$ & $\beta = 0.05$ & $\beta = 0.1$ & $\beta = 0.05$& $\beta = 0.1$ & $\beta = 0.05$& $\beta = 0.1$ & $\beta = 0.05$& $\beta = 0.1$ & $\beta = 0.05$ \\ \midrule
CLIP & - & 77.32 & 77.32 & 64.84 & 64.84 & 88.41 & 88.41 & 90.03 & 90.03 & 80.15 & 80.15 \\ 
PromptFL~\cite{guo2023promptfl} & FPL & 77.73 & 77.76 & 69.13 & 70.08 & 88.86 & 89.35 & 90.04 & 90.38 & 81.44 & 81.89 \\ 
PromptFL~\cite{guo2023promptfl} & CPL & 78.78 & 78.61 & 68.50 & 70.18 & 89.31 & 88.67 & 90.96 & 90.02 & 81.88 & 81.87 \\
Promptprox~\cite{li2020federated} & FPL & 78.00 & 78.00 & 68.25 & 70.32 & 89.21 & 88.73 & 90.61 & 90.72 & 81.52 & 81.94 \\
Promptprox~\cite{li2020federated} & CPL & 78.39 & 75.28 & 68.57 & 60.22 & 89.23 & 86.42 & 90.23 & 87.84 & 81.61 & 77.44 \\
pFedPrompt~\cite{guo2023pfedprompt} & FPL & 80.44 & 79.07 & 69.95 & 69.46 & 89.44 & \textbf{90.59} & \underline{91.44} & \underline{92.23} & \underline{82.82} & \underline{82.84} \\
pFedPrompt~\cite{guo2023pfedprompt} & CPL & 78.78 & 77.34 & \underline{70.64} & \underline{70.42} & 88.64 & \underline{89.77} & 90.07 & 91.07 & 82.03 & 82.15 \\
FedOPT~\cite{li2024global} & FPL & 73.47 & 73.16 & 58.68 & 63.06 & 76.69 & 83.42 & 80.35 & 82.30 & 72.30 & 75.49 \\
FedOPT~\cite{li2024global} & CPL & 77.46 & 76.74 & 66.20 & 67.33 & \textbf{89.88} & 89.51 & 90.98 & 92.21 & 81.13 & 81.45 \\
FedCoPL~\cite{guo2025cooperative} & CoPL & \textbf{81.21} & \textbf{81.66} & 68.89 & 68.76 & 88.42 & 87.65 & \textbf{91.99} & 91.88 & 82.62 & 82.49 \\ \cmidrule{1-12} \rowcolor[gray]{0.9}
\textbf{FedUR} & CoDPL & \underline{81.01} & \underline{80.65} & \textbf{72.12} & \textbf{71.58} & \underline{89.65} & 87.91 & 91.08 & \textbf{92.55} & \textbf{83.47} & \textbf{83.17} \\
\bottomrule 
\end{tabular}
}
\end{table*}

\begin{table*}[t]
\centering
\caption{Ablation study. Accuracies (\%) under Dirichlet-based label skews for four datasets. Conference and Entropy denote confidence-based and entropy-based filters, respectively.}
\label{tab:ablation}
\setlength{\tabcolsep}{8pt}
\resizebox{0.88\textwidth}{!}{
\begin{tabular}{ccccccccc}
\toprule
Data section & Conference & Entropy & Weighted & Negative & DTD & RESISC45 & CUB & UCF101 \\
\midrule
-- & -- & -- & $\checkmark$ & $\checkmark$ & 40.30 & 52.91 & 44.94 & 51.35 \\
$\checkmark$ & -- & -- & $\checkmark$  & $\checkmark$ & 49.20 & 65.32 & 54.73 & 68.73 \\
$\checkmark$ & -- & $\checkmark$ & $\checkmark$ & $\checkmark$ & 60.02 & 74.92 & 54.80 & 71.51 \\
$\checkmark$ & $\checkmark$ & -- & $\checkmark$ & $\checkmark$ & 57.25 & 77.16 & 51.61 & 72.52 \\
$\checkmark$ & $\checkmark$ & $\checkmark$ & -- & -- & 60.89 & 75.76 & 56.09 & 73.20 \\
$\checkmark$ & $\checkmark$ & $\checkmark$ & -- & $\checkmark$ & 62.00 & 78.42 & 56.73 & 73.41 \\
$\checkmark$ & $\checkmark$ & $\checkmark$ & $\checkmark$ & -- & 62.04 & 78.20 & 57.23 & 74.09 \\
$\checkmark$ & $\checkmark$ & $\checkmark$ & $\checkmark$ & $\checkmark$& 62.75 & 78.61 & 57.63 & 74.29 \\
\bottomrule
\end{tabular}}
\end{table*}

\textbf{Results under label and feature skews.} 
Table~\ref{tab:accuracy-label-feature-domain} and Table~\ref{tab:accuracy-label-feature-office} present the performance of all methods under simultaneous label and feature skews on DomainNet and Office-Home, respectively.
In this setting, data within each domain are further partitioned into 3 clients using a Dirichlet distribution ($\beta \in \{0.1, 0.05\}$), introducing both domain-level feature shift and inter-client label imbalance. 
Such compounded heterogeneity significantly complicates reliable pseudo-labeling and federated optimization. 
Most baselines experience noticeable performance degradation and, in some cases, fail to consistently outperform zero-shot CLIP, reflecting their limited capacity to jointly address semantic and distributional discrepancies. 
FedCoPL remains competitive through cooperative pseudo-labeling, yet its performance remains constrained under severe combined skews. 
In contrast, our method, FedUR, consistently achieves the best or near-best performance across all domains on both datasets, attaining the highest average accuracy. 
These results demonstrate that FedUR is more resilient to complex real-world federated scenarios involving both domain and label heterogeneity, highlighting its effectiveness and practical applicability.

\subsection{Ablation Study}
Table~\ref{tab:ablation} presents an ablation study evaluating the contribution of each component in our method under Dirichlet-based label skew across four datasets.
We analyze the impact of four key designs: cooperative sample selection, sample filtering with two different scores, distribution-aware class-wise weighting, and negative regularization.
Starting from a baseline without filtering or global allocation, performance is relatively limited across all datasets. 

Introducing cooperative data selection yields a noticeable improvement, confirming its effectiveness in addressing inter-client label imbalance. 
Adding confidence-based or entropy-based filtering further enhances pseudo-label reliability, with entropy-based filtering providing larger gains due to its stronger ability to discard ambiguous samples.
The weighted component reweights the training loss according to the pseudo-label distribution, enabling better class balancing during optimization. 
The negative component introduces negative supervision by selecting the lowest-confidence classes per sample and minimizing their outputs, which sharpens discriminability and reduces ambiguity.
When all components are jointly enabled, the model achieves the best performance across all benchmarks. 
This demonstrates that cooperative data selection, reliable sample selection, distribution-aware class-wise weighting, and negative regularization are complementary, forming a cohesive and effective design for robust unsupervised federated learning.

\begin{table*}[t]
\setlength{\tabcolsep}{8pt}
\centering
\caption{Performance (\%) of combining our method with baseline methods under Dirichlet-based label skews ($\beta=0.1$).}
\label{tab:accuracy-pl}
\resizebox{0.8\textwidth}{!}{ 
\begin{tabular}{lcllllll}
\toprule
{Method} & \textbf{PL} & {{DTD}} & {{RESISC45}} & {{CUB}} & {{UCF101}} & {{CIFAR10}} \\ \midrule
PromptFL~\cite{guo2023promptfl} & CPL & 44.84 & 62.52 & 48.72 & 63.86 & 88.90 \\
PromptFL~\cite{guo2023promptfl} & \textbf{CoDPL} & 48.77 ({\color{red}+3.93}) & 64.35 ({\color{red}+1.83}) & 49.15 ({\color{red}+0.43}) & 65.71 ({\color{red}+1.85}) & 91.82 ({\color{red}+2.92}) \\
Promptprox~\cite{li2020federated} & CPL & 43.51 & 59.99 & 49.25 & 64.55 & 90.44 \\
Promptprox~\cite{li2020federated} & \textbf{CoDPL} & 48.09 ({\color{red}+4.58}) & 62.73 ({\color{red}+2.74}) & 49.80 ({\color{red}+0.55}) & 66.14 ({\color{red}+1.59}) & 91.91 ({\color{red}+1.47}) \\
pFedPrompt~\cite{guo2023pfedprompt} & FPL & 44.56 & 65.95 & 48.48 & 64.37 & 87.45 \\
pFedPrompt~\cite{guo2023pfedprompt} & \textbf{CoDPL} & 49.67 ({\color{red}+5.11}) & 68.56 ({\color{red}+2.61}) & 52.92 ({\color{red}+4.44}) & 68.23 ({\color{red}+3.86}) & 91.97 ({\color{red}+4.52}) \\
FedOPT~\cite{li2024global} & FPL & 51.97 & 57.76 & 47.99 & 61.83 & 91.32 \\
FedOPT~\cite{li2024global} & \textbf{CoDPL} & 60.07 ({\color{red}+8.10}) & 78.03 ({\color{red}+20.27}) & 56.49 ({\color{red}+8.50}) & 70.07 ({\color{red}+8.24}) & 94.75 ({\color{red}+3.43}) \\
\bottomrule
\end{tabular}
}
\end{table*}

\begin{figure*}[t]
  \centering
  \includegraphics[width=\textwidth]{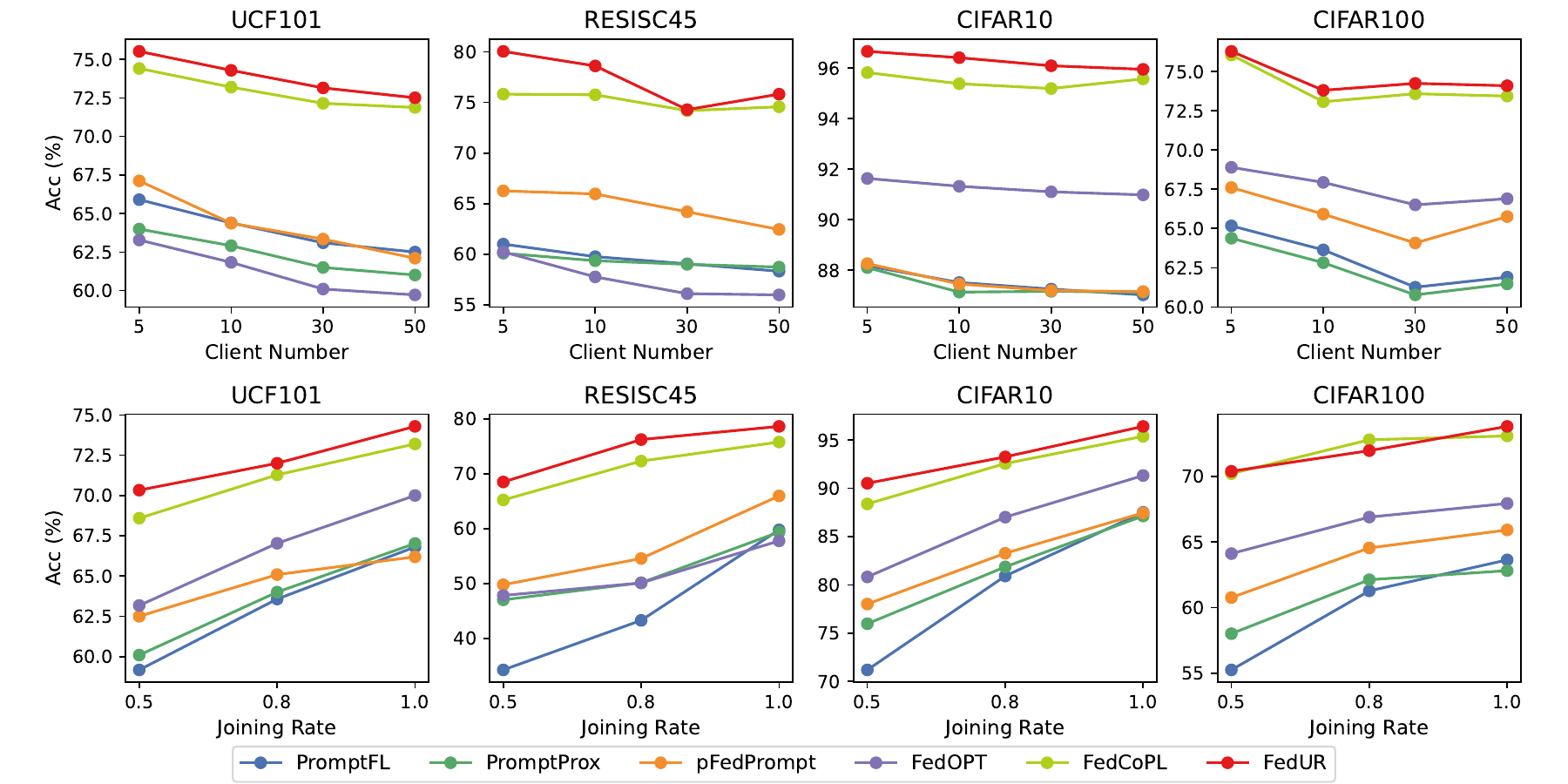}
  \caption{Results with various client numbers and different client joining rates under Dirichlet-based label skews ($\beta=0.1$).}
  \label{fig:client}
\end{figure*}

\subsection{Analysis}
\textbf{Comparison of different pseudo-labeling methods.} 
We evaluate our enhanced pseudo-labeling strategy against widely adopted baselines, including FPL~\cite{menghini2023enhancing} and CPL~\cite{zhang2024candidate}. 
As shown in Tables~\ref{tab:accuracy-dirichlet} and~\ref{tab:accuracy-quantity}, the performance of FPL and CPL varies across datasets: CPL generally outperforms on fine-grained datasets such as DTD and CUB, while FPL is more competitive on CIFAR. 
This variability suggests that existing pseudo-labeling strategies are highly sensitive to both data characteristics and label skew across clients.

Building upon cooperative data selection, we introduce two complementary enhancements to local training: (1) a weighted classification loss that adjusts class weights according to the pseudo-label distribution to mitigate imbalance, and (2) a negative regularization term that imposes explicit negative supervision by suppressing predictions on low-confidence classes. 
We integrate this enhanced pseudo-labeling pipeline into several state-of-the-art supervised FL frameworks by directly replacing their original pseudo-labeling components. 
As reported in Table~\ref{tab:accuracy-pl}, our approach consistently improves performance across all baselines and datasets, yielding substantial gains of up to 20.27\% on RESISC45. 
These results demonstrate that the combination of cooperative pseudo-labeling, distribution-aware weighting, and negative regularization provides a robust and effective solution for federated unsupervised recognition.

\textbf{Results under different client numbers.}
We evaluate the scalability of our method by varying the number of clients on UCF101, RESISC45, CIFAR10, and CIFAR100 under Dirichlet-based label skew ($\beta=0.1$), with results shown in the top row of Fig.~\ref{fig:client}. 
As the number of clients increases, all baseline methods experience a noticeable performance decline, which is a consequence of heightened data heterogeneity and the increasing difficulty of reliable pseudo-labeling across more distributed clients. 
Our method, however, consistently achieves the highest accuracy across all client configurations and datasets, exhibiting significantly slower degradation. 
These results demonstrate that our approach is more resilient to federation scale, maintaining effective learning even as the number of participating clients grows substantially.

\textbf{Impact of client joining rates.}
We further investigate the sensitivity of our method to varying client joining rates, with results shown in the bottom row of Fig.~\ref{fig:client}. 
As expected, reducing the participation rate leads to noticeable performance degradation across all methods, since fewer participating clients limit information aggregation and increase the variance of global updates. 
This effect is more pronounced for baseline methods, which rely heavily on sufficient client participation to stabilize pseudo-label quality and federated optimization. 
On challenging datasets such as UCF101 and RESISC45, baselines experience substantial accuracy drops under lower joining rates, reflecting their limited robustness to partial participation.

In contrast, our method consistently achieves the best performance across all joining rates and datasets. 
Moreover, the performance gap between our method and the baselines becomes larger at lower participation levels, demonstrating that our approach can effectively mitigate the adverse effects of sparse and fluctuating client availability. 
These results confirm that our method, FedUR, is well-suited for realistic federated learning scenarios, where client participation is often intermittent and cannot be guaranteed in every communication round.

\begin{table}[t]
\centering
\caption{Experimental results (\%) using CLIP ViT-B/16 as backbone under Dirichlet-based label skew ($\beta=0.1$).}
\label{tab:accuracy-backbone}
\resizebox{0.45\textwidth}{!}{ 
\begin{tabular}{lccccc}
\toprule
{Method} & \textbf{PL} & {{DTD}} & {{RESISC45}} & {{CUB}} & {{UCF101}}  \\ \midrule
Zero-shot CLIP & - & 42.87 & 56.61 & 55.16 & 65.13 \\
PromptFL~\cite{guo2023promptfl} & FPL & 44.36 & 61.36 & 51.95 & 64.60 \\
PromptProx~\cite{li2020federated} & FPL & 47.12 & 61.01 & 49.44 & 65.31 \\
pFedPrompt~\cite{guo2023pfedprompt} & FPL & 47.65 & 63.18 & 50.19 & 65.19 \\
FedOPT~\cite{li2024global} & FPL  & 46.52 & 58.94 & 49.67 & 65.68  \\ 
FedCoPL & CoPL & 56.83 & 77.50 & 59.78 & 78.29 \\ \rowcolor[gray]{0.9}
FedUR & CoDPL & \textbf{59.69} & \textbf{81.06} & \textbf{59.94} & \textbf{78.61} \\ 
\bottomrule
\end{tabular}
}
\end{table}

\textbf{Results under different image encoder backbones.}
We further evaluate the robustness of our method across different CLIP image encoder architectures. 
Table~\ref{tab:accuracy-backbone} reports the performance of all methods using ViT-B/16 under Dirichlet-based label skew ($\beta=0.1$). 
While ViT-B/16 typically delivers stronger zero-shot performance than ViT-B/32, its adaptation within federated settings with noisy pseudo-labels is more challenging, as it exhibits increased sensitivity to label noise and data heterogeneity. 
Nevertheless, our method consistently outperforms zero-shot CLIP and all baselines across every dataset. 
In contrast, several baseline methods suffer more pronounced performance degradation with ViT-B/16, revealing their limited capacity to effectively utilize stronger visual representations under federated constraints. 
These results confirm that the effectiveness of our approach is not dependent on a specific backbone architecture, underscoring its robustness and practical suitability for real-world federated unsupervised learning deployments.

\section{Conclusion}
In this paper, we extend federated unsupervised learning to classification tasks using CLIP for the first time. 
In this setting, clients with pre-trained CLIP models and unlabeled data collaborate in training to enhance performance without data sharing. 
To address the challenges of CLIP's inherent bias and potential label skew among clients, we propose FedUR, which includes a cooperative data selection strategy, a dual pseudo-labeling mechanism, and a personalized prompt aggregation protocol. 
Specifically, each client estimates and uploads its pseudo-label distribution, and the server adjusts and redistributes them to mitigate global imbalance across categories. 
Moreover, the aggregation protocol aggregates only visual prompts on the server to improve global performance, while textual prompts are kept locally for better personalization by each client. 
Extensive experimental results demonstrate that FedUR consistently outperforms baseline methods across various degrees of label skew, feature skew, and their combination. 
In future work, we will conduct a theoretical analysis of FedUR, including convergence, privacy, fairness, and other pertinent considerations.

\section*{Data Availability}
All datasets used in this study are open-access and have been cited in the paper. The code of this work will be available in the GitHub repository, \url{https://github.com/tim-learn/FedUR/}.

\bibliographystyle{unsrt}
\bibliography{sn-bibliography}

\end{document}